\documentclass[runningheads]{llncs}
\usepackage{textcomp}
\usepackage[utf8]{inputenc}
\usepackage{newunicodechar}
\usepackage{graphicx}
\usepackage{subfig}
\usepackage[T1]{fontenc}
\usepackage{lmodern}
\graphicspath{ {images/} }
\usepackage[ruled,vlined,linesnumbered]{algorithm2e}
\usepackage[outdir=./]{epstopdf}
 \usepackage{lipsum}
 \usepackage{algpseudocode}
\usepackage{amssymb}

\begin{document}
\title{A CP-Net based Qualitative Composition Approach for an IaaS Provider}
%
%
\author{ Sheik Mohammad Mostakim Fattah \and
Athman Bouguettaya \and
Sajib Mistry}
\authorrunning{S. Fattah et al.}
%
\institute{School of Information Technologies, University of Sydney, Sydney, Australia
\email{sfat5243@uni.sydney.edu.au, \{athman.bouguettaya,sajib.mistry\}@sydney.edu.au}}
\maketitle              
\begin{abstract}
We propose a novel CP-Net based composition approach to qualitatively select an optimal set of consumers for an IaaS provider. The IaaS provider’s and consumers’ qualitative preferences are captured using CP-Nets. We propose a CP-Net composability model using the semantic congruence property of a qualitative composition. A  greedy-based and a heuristic-based consumer selection approaches are proposed that effectively reduce the search space of candidate consumers in the composition. Experimental results prove the feasibility of the proposed composition approach. 

\end{abstract}
\section{Introduction}
Infrastructure-as-a-Service (IaaS) model is a cloud service delivery model where computational resources are usually delivered as Virtual Machines (VMs) to cloud consumers \cite{armbrust2010view,mistry2017metaheuristic}. The functional properties of an IaaS service are usually CPU, storage, and memory \cite{chaisiri2012optimization}. The Non-functional properties (e.g., availability, throughput, and price) are usually attached with VMs or IaaS services as Quality of Services (QoS). IaaS services are generally configured based on the functional and non-functional requirements of consumers. For example, Amazon EC2 IaaS provider has different types of VMs (e.g., CPU-intensive, Memory-intensive, and Network-intensive) that are targeted for different types of consumers (e.g., individuals, small enterprises, and large organizations).

The long-term IaaS composition is a topical research issue \cite{mistry2017probabilistic}.  The composition from an IaaS provider's perspective is defined as the selection of a set of optimal consumer requests \cite{mistry2016qualitative}. An effective IaaS composition achieves the economic expectations, i.e., revenue and profit maximization of the provider. The IaaS composition ensures the optimal utilization of available computing resources for an IaaS provider. The selection of optimal consumer requests is essential to achieve the IaaS composition.  For example, selecting service requests from a group of small enterprises may be more profitable than the single service request from a large organization due to the scale of economy.

We focus on the qualitative IaaS composition, i.e., selecting the optimal consumer requests according to the qualitative preferences of the provider. Qualitative preference models are effective tools for the selection where there exists uncertainties or incomplete information. The service requirements of future consumers are uncertain and probabilistic in nature \cite{balke2003towards,liu2013ev,mistry2015predicting}. A provider's preference may change with the requirements of the consumers. The dynamic business environment may also trigger a change in the provider’s qualitative preferences. For example, the provider may observe a very high demand for Network-intensive services in the Christmas or holiday period. The provider may prefer to compose Network-intensive services than CPU-intensive services to increase its revenue.

IaaS consumers’ requirements can be represented in a natural and intuitive manner using qualitative approaches \cite{wang2017measuring}. Qualitative models provide the necessary tool to select appropriate providers where quantitative models are not applicable. A consumer requires to explicitly indicate the exact values of the functional and non-functional properties of a service in a quantitative model. It may not be possible to find providers that can meet the exact requirements of the consumers. For example, a consumer may require 10 units of CPU and 20 units of memory at 20 dollars/month at the level of 100\% availability. Such requirements may not be exactly fulfilled by any service provider. In contrast, qualitative preferences can be expressed by comparison. For instance, a content provider (IaaS consumer) prefers a “high” network bandwidth to a “low” network bandwidth. The content provider may also specify conditional preferences. For example, if the price of network bandwidth is very low, a “high” network bandwidth is preferred over a “low” network bandwidth. These qualitative preferences are used to select suitable providers for consumers.

\textit{The IaaS composition problem} is modeled in both quantitative and qualitative approaches   \cite{mistry2017metaheuristic,mistry2017probabilistic,ye2012qos,ye2016long,li2011semantic}. The quantitative approaches do not consider the qualitative preferences of the provider. The composition of requests is transformed into an optimization problem in quantitative approaches. The proposed approaches (e.g., metaheuristic optimization and integer programming) are not applicable in the qualitative IaaS composition. A heuristic based sequential optimization approach is proposed in the qualitative IaaS composition \cite{mistry2016qualitative}. This approach considers quantitative requirements of the consumers and matches them with the qualitative preferences of the provider. To the best of our knowledge, existing composition approaches are not applicable where both the provider and consumer have qualitative preferences.

\textit{We propose a Conditional Preference Network (CP-Net) based qualitative composition approach for an IaaS provider}. We represent qualitative conditional preferences using CP-Nets. The CP-Net is a very effective tool to represent and reason with qualitative conditional preferences under ceteris paribus (“everything else being equal”) semantics. A CP-Net creates a directed graph where each node is an attribute of a service preference. The edge between nodes defines the priority among service preferences. The rank of service preferences is generated by traversing the graph. We assume that the CP-Nets of the provider and the consumers are provided for simplicity. \textit{Our target is to select the optimal composition of consumers' CP-Nets that has the highest similarity measure with the CP-Net of the provider.}

We propose a CP-Net based qualitative composition approach for an IaaS provider. First, we propose a novel CP-Net composability model to compose CP-Nets of multiple consumers using the semantic congruence property of a qualitative composition. Next, we propose a similarity checking mechanism between CP-Nets using the \textit{coefficient of correlations}. It directs us to apply the brute-force approach where all possible composition of consumers’ CP-Nets is considered to select the optimal composition. The brute-force approach is not a practical solution for composing a large set of consumer's CP-Nets due to its exponential runtime. we propose a heuristic-based algorithm and a greedy algorithm that reduces search space for compositions. The key contribution of our research is summarized below:

\begin{itemize}
\item A CP-Net composability model for the qualitative composition of IaaS consumers using the semantic congruence property.
\item A qualitative similarity measure approach using the correlation coefficient.
\item A heuristic-based and a greedy-based consumer selection approaches to reduce the search space using semantic similarities between the provider’s and consumer’s qualitative preferences.
\end{itemize}

\section{Motivation Scenario}
\begin{figure}[!t]
\vspace{-10mm}
   \centerline{
  \subfloat[]{\includegraphics[scale=0.35]{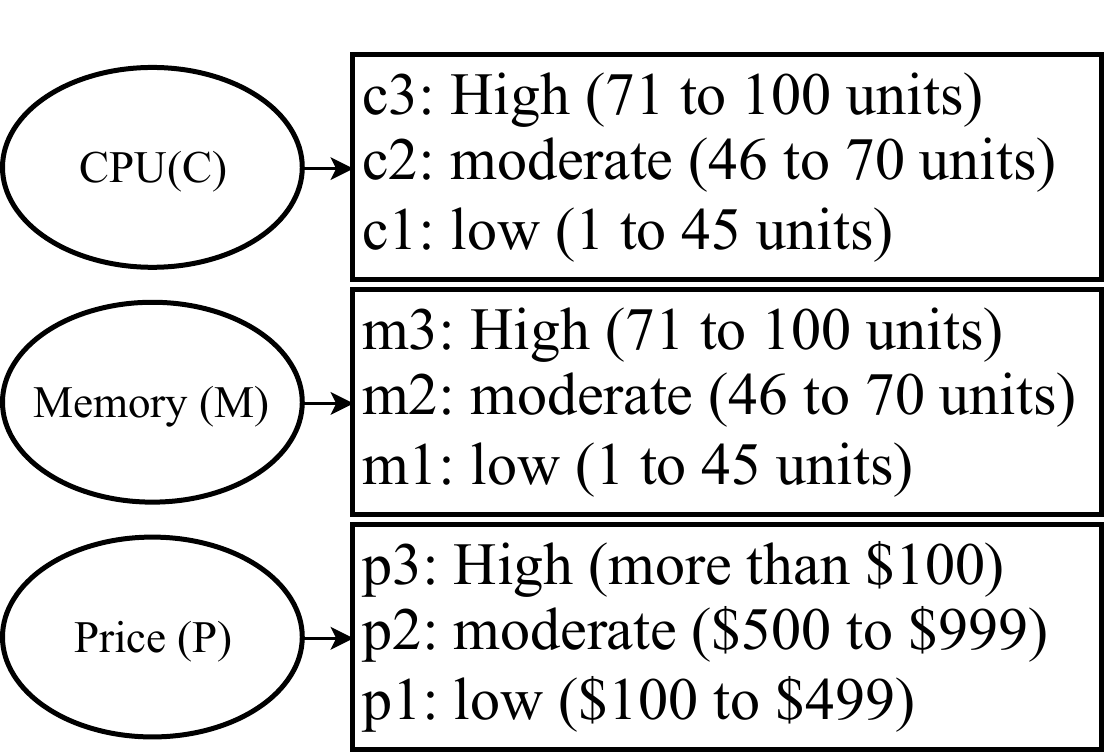} \label{a}}
  \hfil
  \subfloat[]{\includegraphics[scale=0.35]{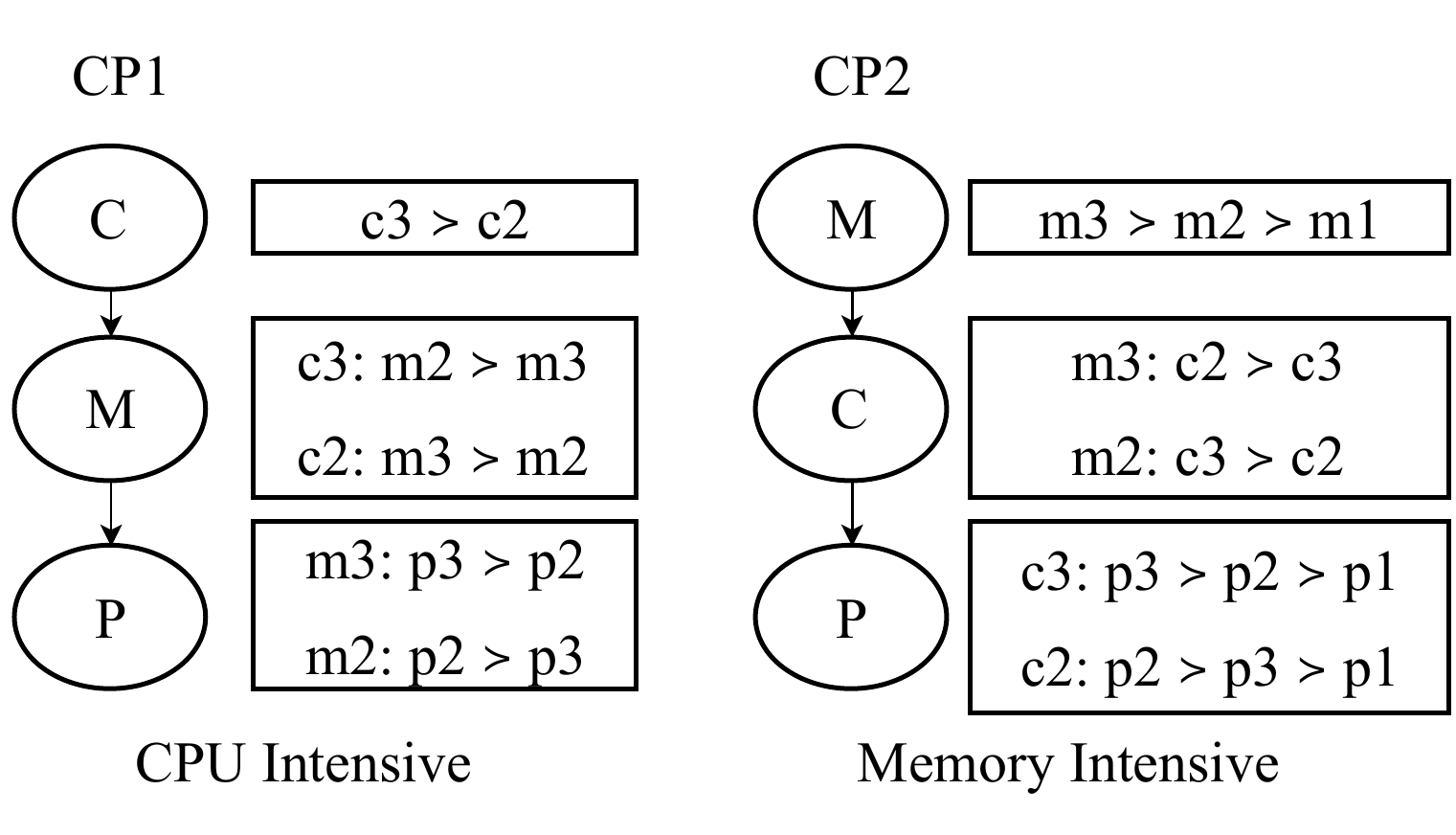} \label{b}}}
  
 \centerline{
  \subfloat[]{\includegraphics[scale=0.35]{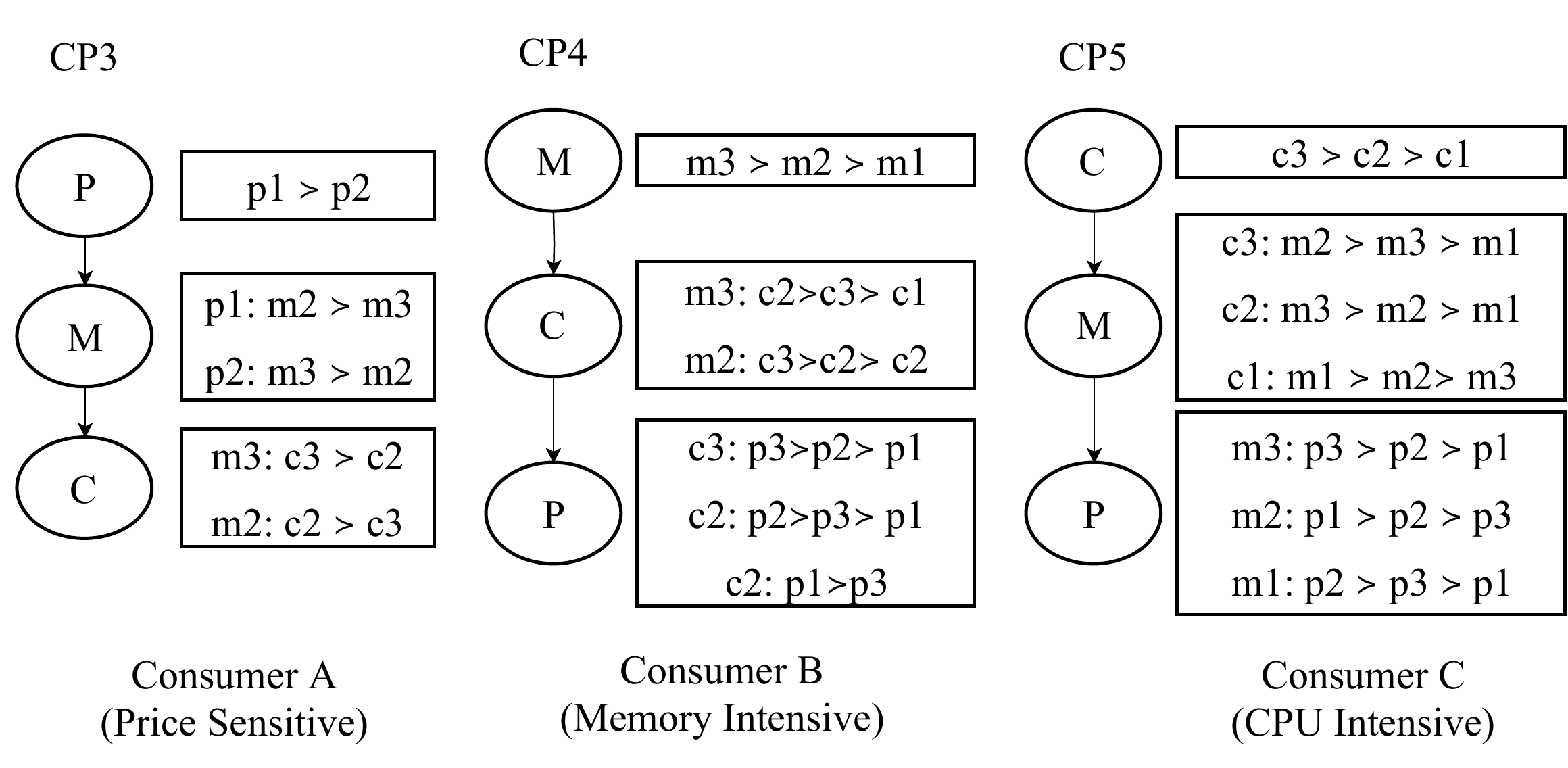} \label{c}}
  }
  
  \caption{(a) Semantic Table for Service Attributes (b) CP-Nets of an IaaS Provider (c) CP-Nets of consumers  }
  \label{fig_mot}
  \vspace{-5mm}
\end{figure}

Let us assume an IaaS provider offers VM services based on a fixed set of computational resources for a specific period of time. Its resource capacity is up to 100 virtual CPU units and 100 memory units. For simplicity, we consider “price” as the only QoS in a VM and we omit “network bandwidth (NB)” functionality from a VM. We also assume that both the consumers’ qualitative requirements and the providers’ qualitative preferences on CPU, memory, and price are following the same semantic levels in Figure \ref{fig_mot}(a). Three levels of semantics, i.e., high, moderate, and low for each attribute are specified in the semantic table in Figure \ref{fig_mot}(a). The IaaS provider builds different types of strategies of service provisions. As there exist uncertainties on future consumers’ requirements, it builds the strategies using qualitative models. We assume that the provider receives qualitative requests from three consumers. The target is to select an optimal set of consumer requests that matches with its preferred ways to service provisions.

We consider a “CPU-intensive” and a “memory-intensive” service provisions strategies for the provider. As the CP-Net provides an effective way to represent conditional qualitative preferences, we represent the “CPU-intensive” strategy as CP1 and the “memory-intensive” strategy as CP2 in Figure \ref{fig_mot}(b). The “CPU-intensive” strategy is to offer CPU intensive services at relatively moderate prices to attract consumers with CPU intensive requests. Therefore, the CPU is the most important attribute in the dependency graph of CP1 followed by memory and price. The arc from “CPU” to “memory” implies that the provision of “memory” levels in a VM depends on the selection of “CPU” levels. The preference of CPU provisions is expressed as \(c3 \succ c2\) in the CPT of CPU. It implies that the provider does not want to offer the “low” CPU services. The choice of CPU levels decides the choices of memory levels in CP1. The provider prefers the “moderate” memory to the “high” memory if the “high” CPU is chosen in the service provision according to CP1. If the choice of CPU is “moderate”, the provider prefers the “high” memory to the “moderate” memory. Based on the selection of memory levels, the provider selects the price levels if the “high” memory is chosen according to the CPT of price in CP1. The provider prefers the “high” price to the “moderate” price. However, the provider prefers the “moderate” price to the “high” price for the “moderate” memory units. Similarly, CP2 represents the “memory-intensive” strategy where memory is the most important attribute in the dependency graph of CP2 followed by CPU and price.

The requirements of consumer A, B, and C are captured by CP3, CP4, and CP5 respectively in Figure \ref{fig_mot}(c). Consumer A prefers “price-sensitive” services. The price is the most important attribute in the dependency graph of CP3 followed by memory and CPU. The Consumer A mentions its inability to consume “high” priced (P3) services due to budget constraints and prefers the “low” price to “moderate” priced services i.e., \(p1 \succ p2\) according to the CPT of price in CP3. Its memory requirements depend on the choice of price. In the CPT of memory in CP3, the “moderate” memory is preferred over the “high” memory in the required service if the “low” price is already chosen. If the “moderate” price is chosen, it sets a higher priority to “high” memory than the “moderate” memory. Preferences of CPU requirements are defined based on the choice of memory in a similar way. The consumer B has “memory-intensive” qualitative requirements. Therefore, memory is the root attribute in CP4. The choice CPU depends on the choice of memory and the choice of price depends on the choice of CPU in CP4. The consumer C defines “CPU-intensive” preferences in a similar way.

The IaaS provider can select the optimal composition of consumer’s requests, i.e., CP-Nets from three consumers in \(2^3 -1\) = 7 possible ways. Multiple consumer’s CP-Nets may not be composable for their semantic differences. For example, the “Price-sensitive” requirement (CP3) of consumer A prefers to receive the “low” memory level services if the price is at a “low” level. The “memory-sensitive” requirement (CP4) of consumer B prefers to receive the “high” memory level services even at a “high” level. Hence, CP3 and CP4 are not composable as they create semantic ambiguity in the composition. We propose a composability model to determine whether CP-Nets of two consumers are composable or not.

Finding all possible compositions of preferences in a brute force approach can be very inefficient. The number of all possible combination of N consumers is (\(2^N -1\)). We propose a heuristic algorithm to reduce the search space of consumers by finding the relative similarity between the consumers’ CP-Net and the provider’s CP-Net. For example, if the provider’s strategy is “CPU-intensive”, i.e., CP1; selecting the “price-sensitive” consumer A in a composition may not yield the optimal result as they are semantically very dissimilar. The search space reduces to 2 consumers (consumer B and C) and the optimal composition is selected from \(2^2 -1\) = 3 compositions out of the 7 possible ways.

\section{CP-Nets for Qualitative IaaS requests and provisions}

A CP-Net is a graphical model to formally represent and reason about qualitative preference relations. A CP-Net consists of a directed dependency graph and conditional preference tables (CPTs). The dependency graph is defined over a set of functional and non-functional attributes \(V=\{X_1,…., X_n\}\).  A child node in a dependency graph depends on a set of direct parent nodes \(Pa(X_i)\). The child node is connected by an arc from \(Pa(X_i)\) to \(X_i\) in the dependency graph. Parent attributes affect the user's preferences over the value of \(X_i\). Each node \(X_i\) in the dependency graph has \(Pa(X_i)\) except for the root nodes.

The CPT of each variable \(X_i\) is defined over the finite, discrete domain \(D(X_i)\) and semantic domains \( S(X_i) \). Each value \(x_n\) in \(D(X_i)\) is mapped into a semantic value in \(S(X_i)\) using a semantic mapping table, \(SemTable(X_n, x_n)\). Figure \ref{fig_mot}(a) is a semantic table that maps 71-100 units of CPU as a “high” CPU value. We only focus on the attributes that are compatible with additive operations. Hence, we define \( s_i + s_j = s_k \) for \(S(X_i)\). For example, \(c1+c1=c2\) implies two “low” CPU can be added and generates a “moderate” CPU unit. 

The preference between two values of an attribute \(X_i\) is specified by \(\succ\) for a given value of the paraent attribute \(Pa(X_i)\). A user explicitly defines its preferences over the semantic values of \(X_i\) for each complete outcome on \(Pa(X_i)\). The preferences take the form of total or partial order over \(S(X_i)\). For example, the attributes of CP1 are \(C\), \(M\), and \(P\) with semantic domains containing \(x_i\) if \(X\) is the name of the feature. The preferences statements are as follows: \( c3 \succ c2 \), \( c3: m2 \succ m3\), \(c2: m3 \land m2 \), \(m3 : p3 \succ p2 \), \( m2 : p3 \succ p2\). The statement \(x_1 \succ x_2\) represents the unconditional preference for \(X=x_1\) over \(X=x_2\).

\begin{figure}[!t]
  \centering
    \includegraphics[scale=0.4]{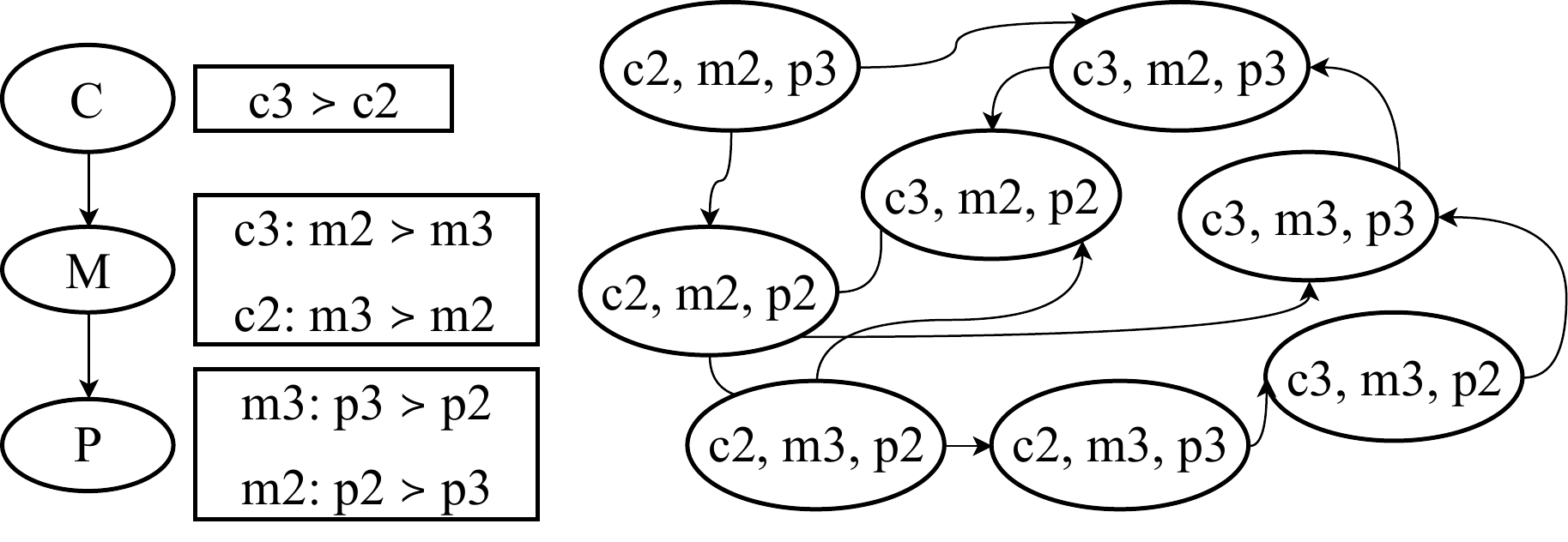}
    \caption{Induced Graph}
  \label{fig_ind}
  \vspace{-5mm}
  \end{figure}

A preference outcome is a combination of values of all attributes of a CP-Net. For example, \(\{c3, m2, p3\}\) and \(\{c2, m2, p1\}\) are two preference outcomes for CP1 denoted by \(o_1\) and \(o_2\). According to the value of attribute \(P\), it can be shown that \(o_1 \succ o_2\) or \(o_1\) dominates \(o_2\). The dominance relationship of two preference outcomes is defined as a \textit{pre-order} between them. Figure \ref{fig_ind} depicts the induced graph \cite{boutilier2004cp} of a CP-Net with all preference outcomes.

\section{Composability Model for Qualitative Preferences}

Let us consider two CP-Nets \(CP_A\) and \(CP_B\). We define the composability of two CP-Nets as \(composable(CP_A, CP_B)\)  = \(\{true,false\}\) to find \(CP_C\) where \(CP_C =  compose(CP_A, CP_B)\). Two CP-Nets are composable if their composition is semantically congruent. \textit{A composition is called semantically congruent if the relative importance order of preference attributes for each consumer is preserved without any ambiguity.} The relative importance order is represented as \((a, b)\) which means \(a\) is preferred over \(b\). Let us consider a consumer who has the relative importance order of attributes as  \((CPU, memory)\). Another consumer has the relative importance order of attributes as \((memory, price)\). If the importance order of their composition is \((CPU, memory, price)\), the composition is considered as semantically congruent. 

\begin{definition}{\textit{Semantic Congruence of a Qualitative Composition}}.  
A composition is called semantically congruent when the importance order of preference attributes for each consumer is preserved without any ambiguity.
\end{definition}

\begin{figure}[!t]
   \centerline{
  \subfloat[]{\includegraphics[scale=0.3]{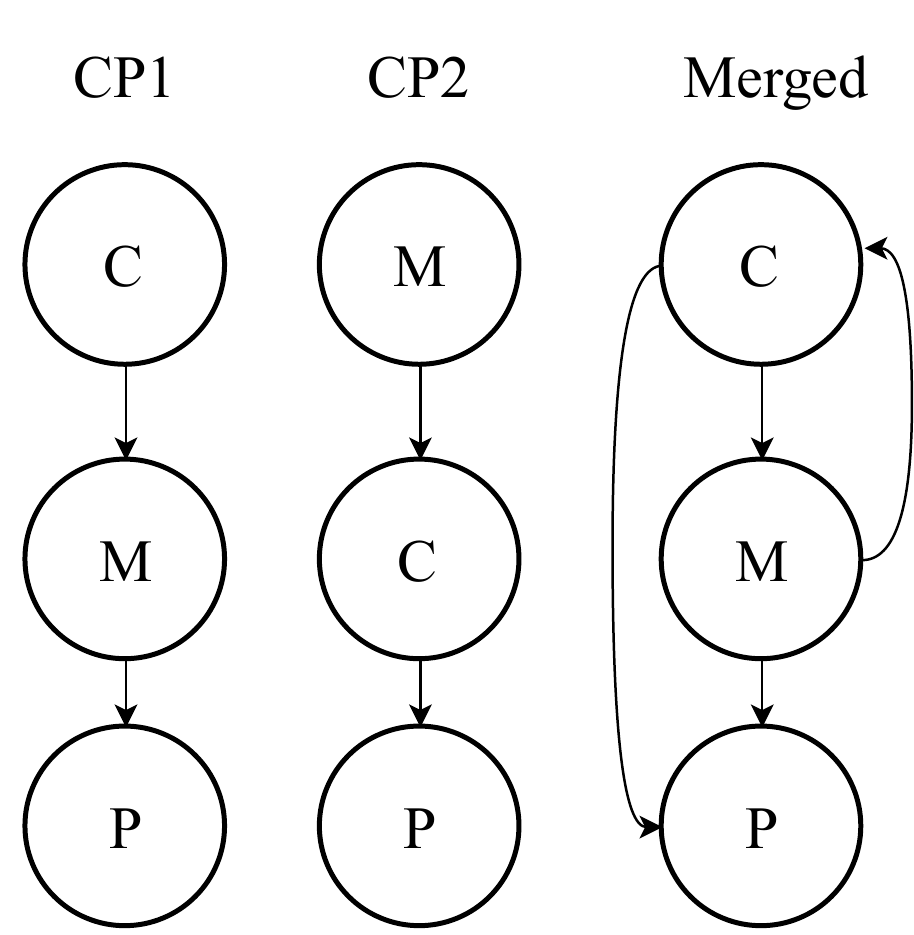} \label{a}}
  \hfil
  \subfloat[]{\includegraphics[scale=0.35]{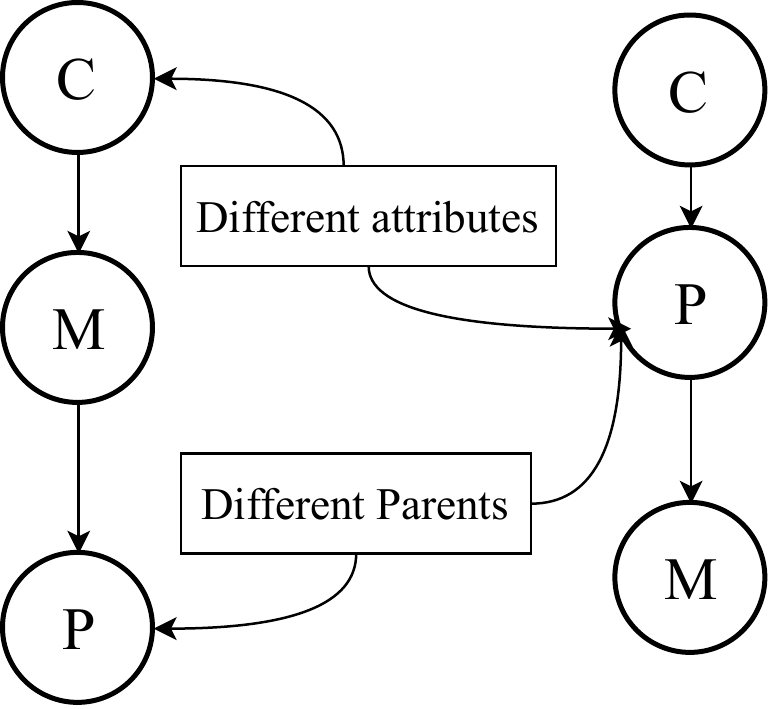} \label{b}}}
  
  \caption{(a) Dependency Graph Composability  (b) CPT Composability  }
  \label{fig_comp}
  \vspace{-5mm}
\end{figure}

The semantic congruence of a composition can be efficiently represented using the directed acyclic graph of CP-Nets. We use semantic congruence property to define the composability of two CP-Nets. A CP-Net contains a directed dependency graph (DDG) and conditional preference table for each node in the graph. We define the composability of the DDG and the CPT to define the composability of CP-Nets. Two DDG are considered to be composable if their combined DDG does not contain any cycle. Figure \ref{fig_comp}(a) shows two DDGs from two different CP-Nets (CP1 and CP2). The CPU is the root of CP1. The memory depends on the CPU and the price depends on the memory in CP1. The DDG of CP2 has the order of memory, CPU, and price. To merge them in a single DDG (i.e., CP12) (Figure \ref{fig_comp}(a)), we create a new DDG where all the attributes (i.e., CPU, memory, and price) are added from both CP-Nets. The next step is to create edges between the attributes. First, we take a pair of attributes (e.g., CPU and price) from the new DDG. If the same pair of attributes has an edge in either DDG (i.e., CP1 or CP2), we add an edge to the new DDG. We run this process for each pair of nodes until we cover all edges from both DDGs. If the resulted DDG contains any cycle, then DDGs are not composable. Two CP-Nets are composable if their dependency graphs and CPTs are composable.

\begin{definition}{\textit{CP-Net Dependency Graph Composability}}.  
Two dependency graphs are composable if their combined dependency graph does not contain any cycle.
\end{definition}
\begin{definition}{\textit{CPT Composability}}.  
Two CPTs from two different CP-Nets are composable if their preference attributes are same and their values depend on the same set of parent nodes.
\end{definition}

Two CPTs are composable if they have two properties. First, both CPTs’ attribute nodes should be the same. In Figure \ref{fig_comp}(b), “CPU” nodes from both DDGs can be valid candidates to be composable. A “CPU” node from one CP-Net and a “memory” node from another CP-Net are not composable. Second, the parent nodes of both nodes should have the same attribute. If two nodes have the same set of parent nodes, then preference statements of both nodes will depend on the same set of attributes. Therefore, the preference statements will be composable. For example, “CPU” nodes of both CP-Nets are unconditional nodes (i.e., no parent) in Figure \ref{fig_comp}(b). However, “price” nodes have different parents (i.e., “memory” and “CPU”). The preference statements will have conditional attributes. Therefore, the CPTs of the "price" nodes are not composable.

\section{Similarity Measurement with Qualitative Preferences}

We assume that the IaaS provider expresses its qualitative preferences using the CP-Net, \(CP_A\). The provider requires to find an optimal set of consumer CP-Nets from \(\{CP3, CP4, CP5\}\) that matches with \(CP_A\). We apply the composability model described in the above section and find the composed CP-Net \(CP_B\) for a set of composable consumers. If \(CP_A\) completely matches with \(CP_B\), we say it is the optimal composition. To compare the composed CP-Net with provider’s CP-Net we need to find a similarity measurement algorithm. The similarity measurement between two CP-Nets can be performed in two ways \cite{wang2009web,wang2017incorporating}. One way is to generate the induced graph of two CP-Nets and compute the number of common edges between two CP-Nets. This similarity can be computed by:

\begin{equation}
  \resizebox{.9\hsize}{!}{$Sim(CP_A : CP_B) = \frac{|\{e : e \ \in \ In(CP_A) \ \land \ e \ \in \ In(CP_B) \}|}{|\{e:e \ \in \ In(CP_A) \  \lor \ e \ \in \ In(CP_B) \}|\ - \ |\{e:e \ \in \ In(CP_A) \ \land \ e \ \in \ In(CP_B) \}|}$}
  \label{eqn:induced_check}
\end{equation}

\noindent where \(In(CP_A)\) and \(In(CP_B)\) denotes the induced graph for \(CP_A\) and \(CP_B\). The edge between two attributes is denoted by \(e\). The equation \ref{eqn:induced_check} computes the ratio between the number of common and total edges between two induced graphs. This method is computationally expensive and not applicable in real time \cite{wang2017measuring}. 

Another way is to compare the CPTs between two CP-Nets using the dependency graphs. This method is only applicable when two CP-Nets share the same dependency graph \cite{wang2017incorporating}. In that case, the similarity between the provider’s and consumers’ CP-Net is calculated by the following equation:

\begin{equation}
  \resizebox{.9\hsize}{!}{$Sim(CP_A:CP_B) = \frac{ \sum_{X_i} \Big(|CPT_A(X_i) \cap CPT_B(X_i)| \times \prod_{X_j \notin Pa(X_i) } |SemTable(X_i)| \Big)}{\sum_{X_i} \Big(|CPT_A(X_i) \cup CPT_B(X_i) | \times \prod_{X_j \notin Pa(X_i) } |SemTable(X_j)| \Big)}$}
  \label{eqn:cpt_check}
\end{equation}

\noindent where \(CPT_A\) and \(CPT_B\) are the conditional preference table of \(CP_A\) and \(CP_B\).  \(Pa(X_i)\) denotes the parent attributes of \(X_i\) and \(SemTable(X_i)\) represents all values that can be assigned into \(X_i\). We assume the composed CP-Net of consumers and the provider's CP-Net have the same dependency graph.  

\section{Heuristic-based Consumer Selection approaches }

The number of composable consumers grows exponentially with the increase in the number of consumers (\(2^n)\). Finding all possible combinations of consumers is inapplicable as it may require a very large time depending on the number of consumers \cite{mistry2017probabilistic}. Our target is to reduce the search space for the consumer selection. We propose a greedy-based and a heuristic based consumer selection algorithm using the similarity between a provider’s and a consumer’s CP-Nets.

\subsection{Greedy-based IaaS Consumer Selection Approach}
We choose the first consumer who has the highest relative similarity with provider’s CP-Net in the greedy selection approach. We iteratively choose the next consumers to achieve the maximum similarity with the provider's CP-Net.  The following steps are performed in the greedy based approach:   

\begin{enumerate}

\item Select a consumer CP-Net that is has the maximum coefficient of correlation with the provider’s CP-Net.
\item Create a new CP-Net based on the difference between the provider’s CP-Net and the selected consumer’s CP-Net. 

\item Find and select a consumer CP-Net who has the maximum correlation coefficient with the new CP-Net.
\item Create a new CP-Net based on the difference between the consumer CP-Net and the new CP-Net.
\item Perform step 3 and 4 until the difference is zero or minimum.   
\end{enumerate}

\subsection{IaaS Consumer Selection based on Correlation Coefficients}

The greedy approach may not always provide the accurate results as it considers only consumers with maximum correlation with the provider’s CP-Net. We proposed a heuristic approach where we find those consumers who have relatively similar CP-Nets with the provider’s CP-Net. Relatively similar CP-Nets are more likely to form a composition that will match the CP-Nets of the provider. Two CP-Nets are relatively similar if (1) they have the same dependency graph (2) Nodes with similar attributes have similar preferences statements in their CPTs. CP1 and CP5 have the same dependency graph in Figure \ref{fig_mot}.  The relative similarity between two preferences statements is measured based on their relative ordering. For example, consider two preferences statements \(c1 \succ c3 \succ c2\) and \(c8 \succ c10 \succ c9\). Although values of the attributes are different, patterns of both statements are same. We consider \(c1 \succ c3 \succ c2\) and \(c8 \succ c10 \succ c9\) are relatively similar. Let us consider two conditional preferences statements \(c1 \land m2 : p1 \succ p2 \succ p3\) and \(c8 \land m10: p6 \succ p7 \succ p8\). The condition of the first statement \(c1 \land m2\) can be fulfilled by the condition of the second statement \(c8 \land m10\). A similar statement can be found in the CPT of the same attribute of the provider’s CP-Net for each statement of the CPT of an attribute from the consumer’s CP-Net. We perform the following steps to find the relative similarity between a consumer’s and the provider’s CP-Net:

\begin{enumerate}
\item  Compare the dependency graph of the provider’s and the consumer’s CP-Net. If the dependency graphs are not the same, the CP-Nets are not similar.  
\item If the dependency graphs are same, find an unconditional node from provider’s CP-Net for each unconditional node of the consumer CP-Net.

\item Compute similarity between the unconditional nodes selected in step 2.
\item Store the similarity measurement in a global variable.
\item Find similar conditional nodes for each attribute from provider’s and consumer’s CP-Nets. 
\item For each preference statement in a CPT of an attribute of the consumer’s CP-Net, find a similar preference statement in the CPT of the same attribute of provider’s CP-Net. The attributes and the conditions of both statements should be also relatively similar. 
\item The similarity between the conditional nodes is computed. Update the total similarity measurement.

\end{enumerate}

\begin{algorithm}[H]
  \DontPrintSemicolon
  \SetAlgoLined
  \SetKwInOut{Input}{Input}\SetKwInOut{Output}{Output}
  \Input{$CP_A$, $CP_B$, $SemanticTable$}
  \Output{Similarity $Sim(CP_A,CP_B)$}
  \BlankLine

  Integer $commonEdges \leftarrow 0 $ \;
  Integer $allEdges \leftarrow 0 $ \;
  $CPT_A \leftarrow $ find all CPT in $CP_A$\;
  $CPT_B \leftarrow $ find all CPT in $CP_B$\;
 
  \BlankLine
  \ForEach{ $X_i$ attribute in $CP_A$}{
    \BlankLine
    $ visitedPreferences \leftarrow \varnothing $ \;
    \ForEach{$P_A$ in $CPT_A[X_i]$}{
      boolean $flag \leftarrow false $ \;
      \ForEach{$P_B$ in $CPT_B[X_i]$ }{
        \If{$P_A$ has similar pattern $P_B$ }{
        $visitedPreferences \leftarrow P_B$ \;
        $flag \leftarrow true$\;
        $commonEdges \leftarrow commonEdges + \prod_{X_j \notin P(X_i) |SemTable(X_j)|} $\;
        $allEdges \leftarrow allEdges + \prod_{X_j \notin P(X_i) |SemTable(X_j)|} $\;

        }
        \If{$!flag$}{
          $allEdges \leftarrow allEdges + \prod_{X_j \notin P(X_i) |SemTable(X_j)|} $\;
        }
      }
    }
    \ForEach{$P_B$ in  $CPT_B[X_i]$}{
      \If{$P_B \notin visitedPreferences$}{
        $allEdges \leftarrow allEdges + \prod_{X_j \notin P(X_i) |SemTable(X_j)|} $\;
      }
    }
  }
  return $Sim(CP_A,CP_B) \leftarrow commonEdges/allEdges $\;
  \caption{Similarity Checking between two CP-Nets}
  \label{alg:sim_alg}

  \end{algorithm}

We propose Algorithm \ref{alg:sim_alg} to find relatively similar consumers based on the  provider’s CP-Nets. Algorithm \ref{alg:sim_alg} calculates the coefficient of correlation between two CP-Nets using equation \ref{eqn:cpt_check}. The algorithm takes two CP-Nets with same dependency graphs. Two variables are defined to calculate the number of common edges and all edges between the CP-Nets (\(commonEdges\) and \(allEdges\)). According to our assumption, both CP-Nets have the same number of attributes. For each attribute, we perform a check if the conditional preferences from both CP-Nets have a similar pattern. When a preference has a similar pattern in both CP-Nets, we update the number of common edges (\(commonEdges\)) and all edges (\(allEdges\)). The preference is added in \(visitedPreferences\). However, if there is no preference from \(CP_A\) is found, the algorithm updates only the number of all edges (\(allEdges\)). \(allEdges\) is updated with every iteration. The relative similarity is calculated by the ratio of the number of common edges (\(commonEdges\)) and the number of all edges (\(allEdges\)).

\section{Experiments and Results}
We have conducted a set of experiments to evaluate the efficiency and the feasibility of the proposed heuristic based composition approach. The heuristic and the greedy approaches are compared with the brute force approach in term of accuracy and time. We conducted the experiment on computers with Intel Core i7 (3.60GHz and 8GB RAM) using Java and Matlab.

\subsection{Simulation Setup}
It is difficult to find the real-world preferences of IaaS consumers. We have generated 20 CP-Nets to represent consumers’ preferences. We have also generated a semantic table for consumers which is a subset of provider’s semantic table. The provider has the entire view of its resource capacity. As the simulation has been performed based on randomly generated CP-Nets, the result varies depending on the type of the CP-Nets. We run the experiment based on Monte Carlo \cite{binder1993monte} simulation method for a conclusive result. We have run the simulation several times for each approach and taken the average accuracy and time for the different size of consumers. Table \ref{tab1} shows the simulation variables and their corresponding values that we have used in the experiment to perform the performance analysis.

\begin{table}[!t]
\vspace{-5mm}
\centering
\caption{Experiment Variables}\label{tab1}
\begin{tabular}{|l|l|}
\hline
{\bfseries Variable Names} & {\bfseries Values}\\
\hline
Simulation Run &  {100} \\
Number of Consumers &  { 2 to 23} \\
Coefficient of Correlation & { 0.15, 0.20, 0.25 }\\
Homogeneous Domain Size &{ 20} \\
\hline
\end{tabular}
\vspace{-5mm}
\end{table}

\subsection{Baseline: The Brute-force Approach}
We generate all combination of the consumers for the brute force approach. For each combination, we compose them using the composability model. The set of composable CP-Nets are composed and compared with the provider’s CP-Net using equation \ref{eqn:induced_check}. A composed CP-Net that has maximum similarity with the provider’s CP-Net is selected. As the brute force approach considers all consumers, it achieves maximum similarity up to 90\% with provider’s CP-Net.

\begin{figure}[!t]
\vspace{-5mm}
   \centerline{
  \subfloat[]{\includegraphics[width=0.5\textwidth]{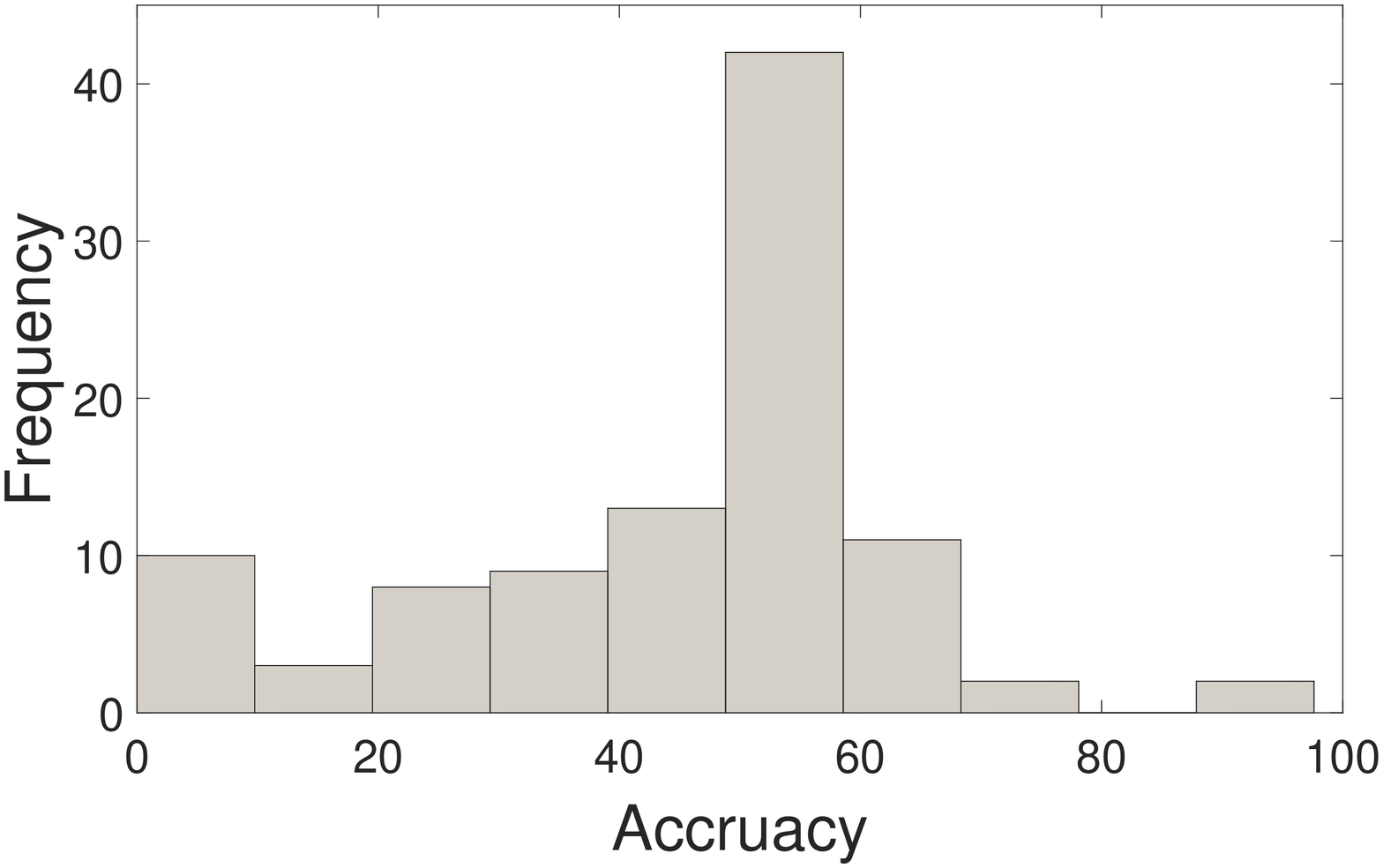} \label{a}}
  \hfil
  \subfloat[]{\includegraphics[width=0.5\textwidth]{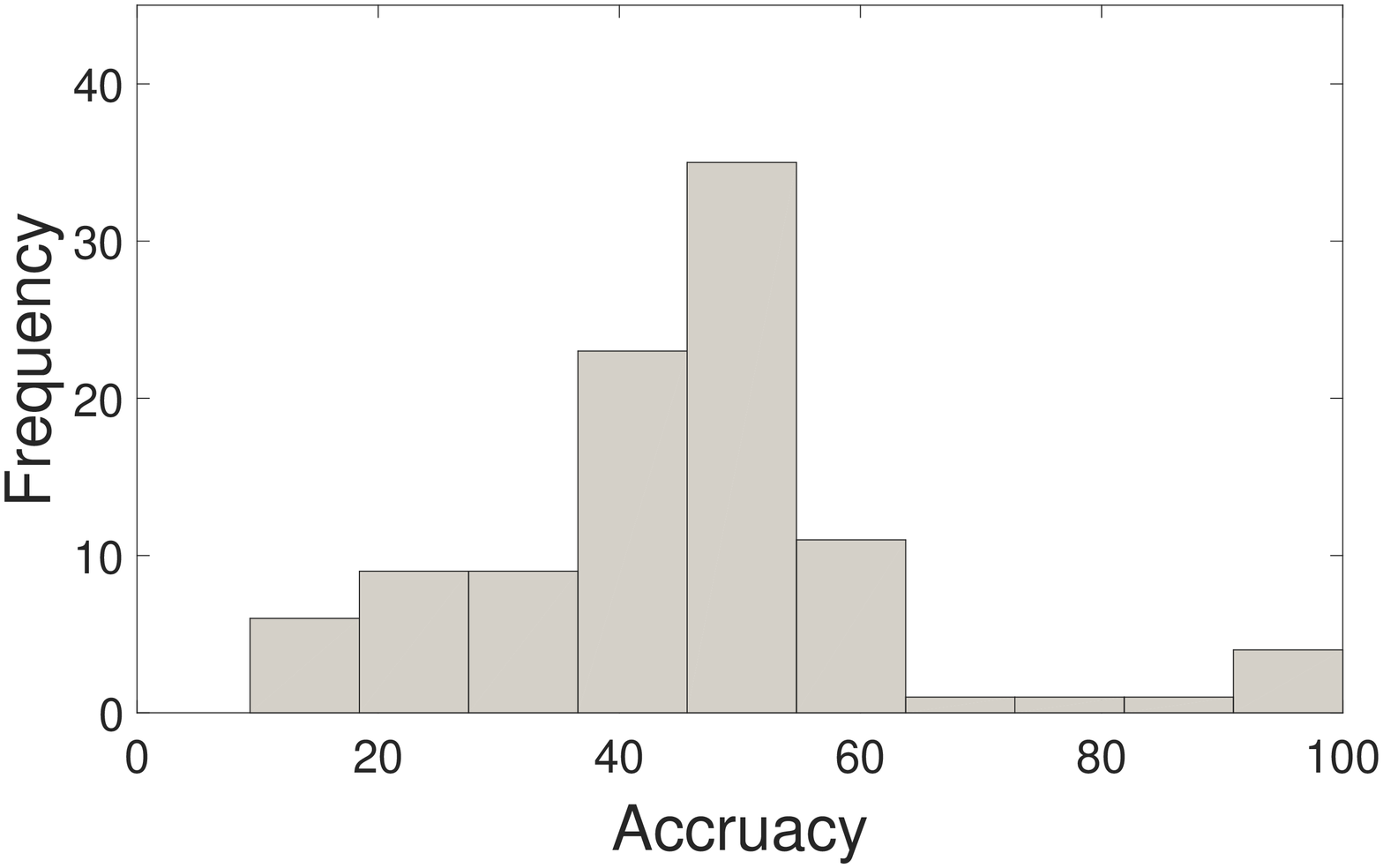} \label{b}}}

  \centerline{
  \subfloat[]{\includegraphics[width=0.5\textwidth]{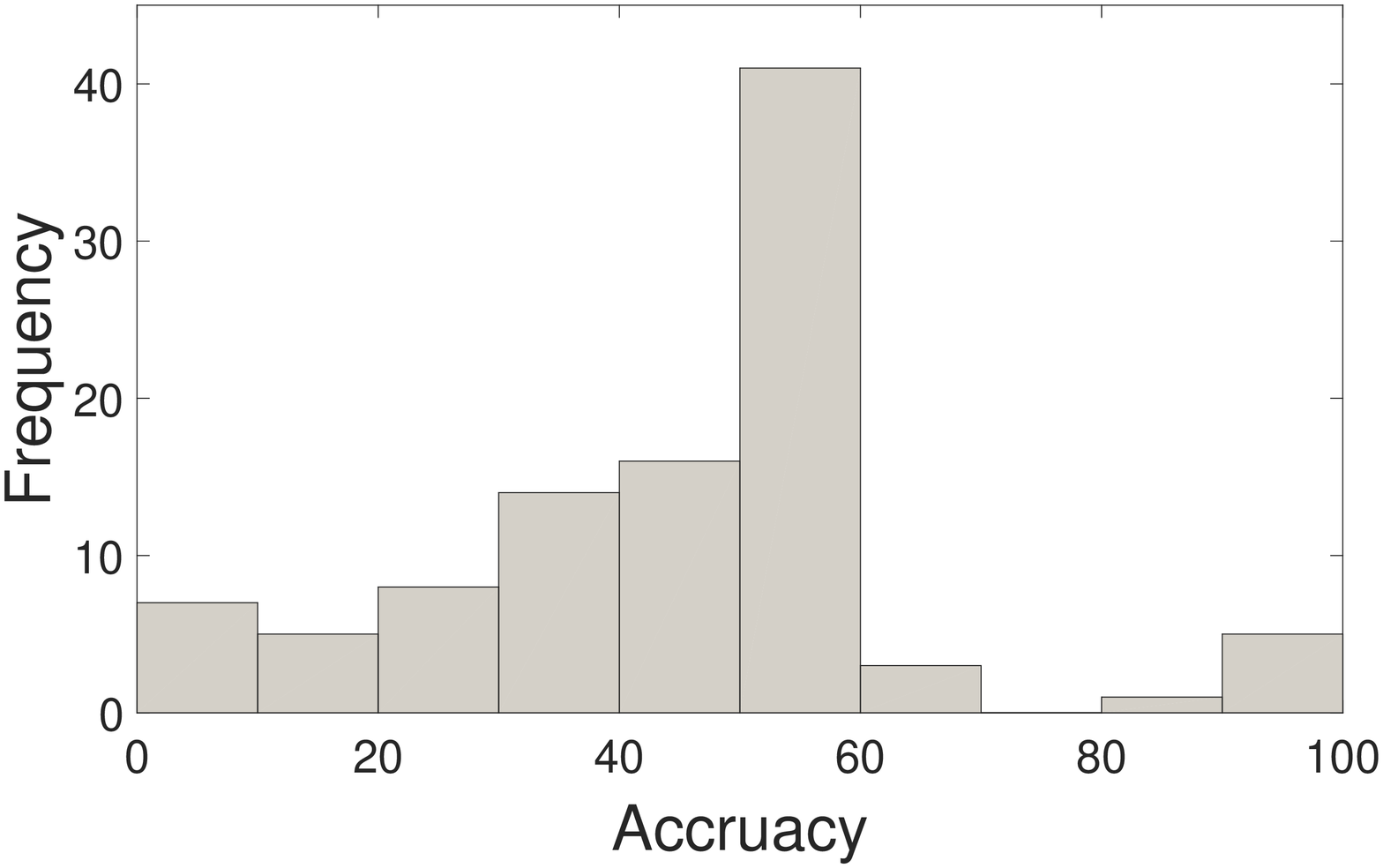} \label{a}}
  \hfil
  \subfloat[]{\includegraphics[width=0.5\textwidth]{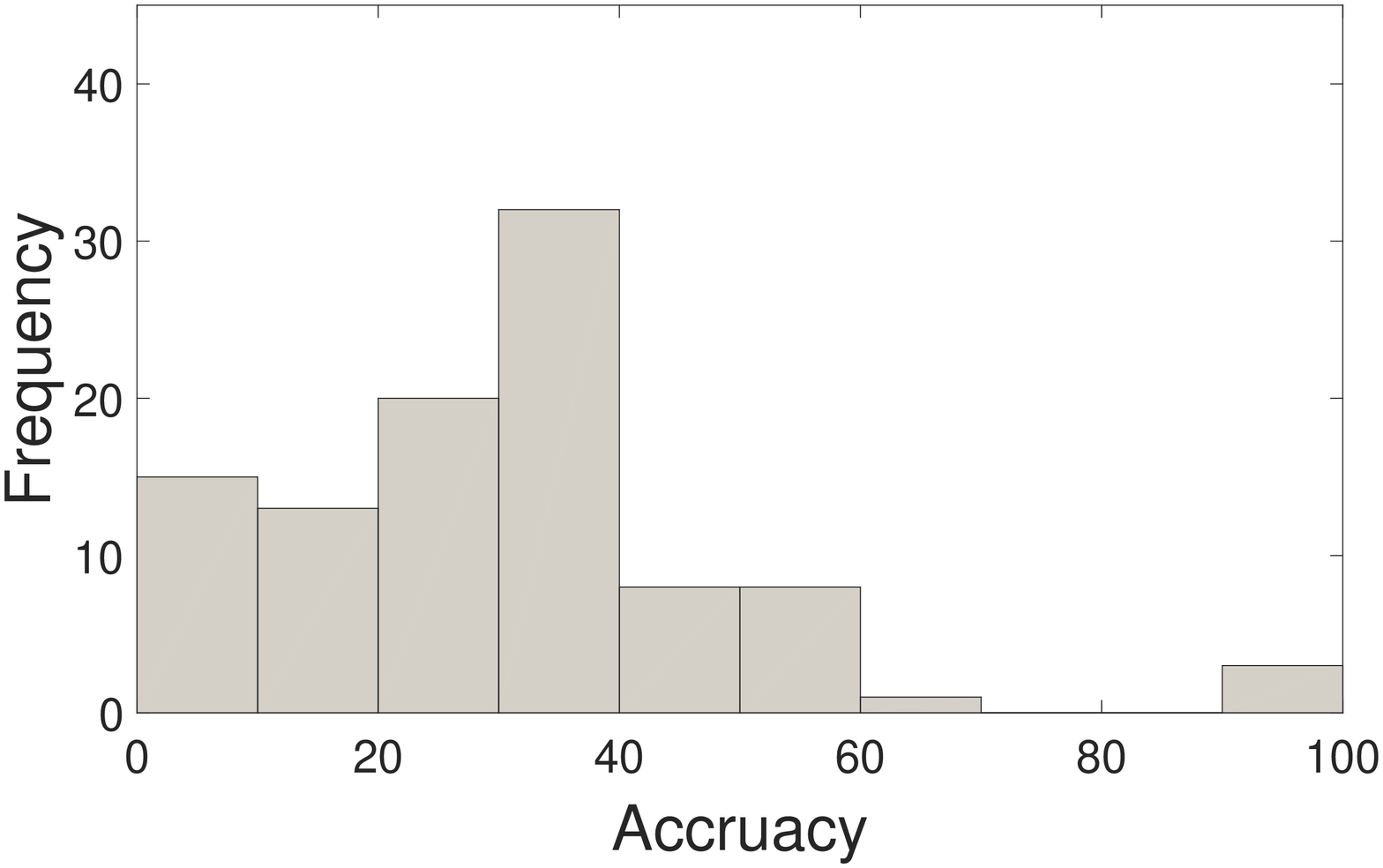} \label{b}}}

   \centerline{
  \subfloat[]{\includegraphics[width=0.5\textwidth]{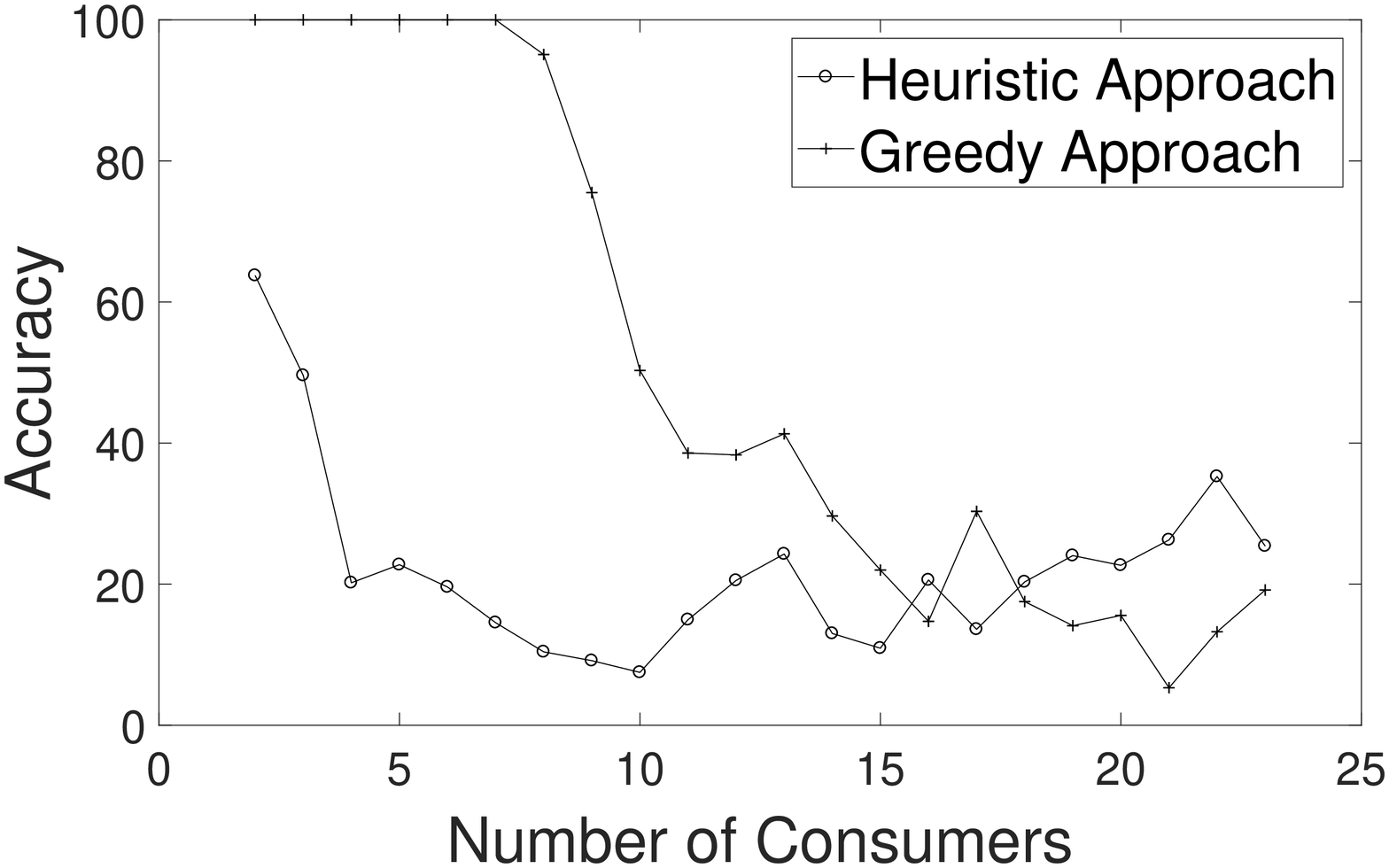} \label{c}}
  \hfil
  \subfloat[]{\includegraphics[width=0.5\textwidth]{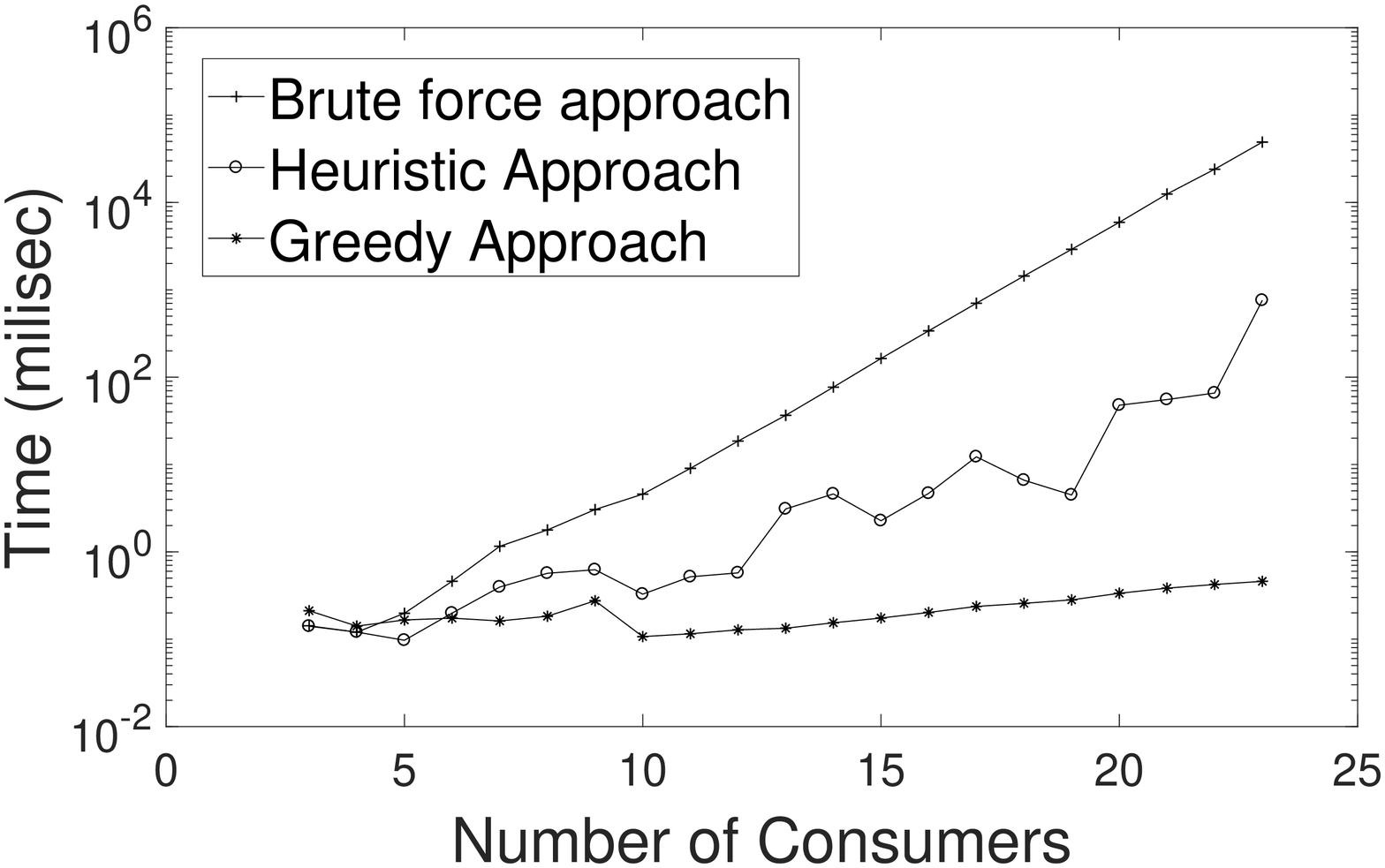} \label{d}}}
  
  \caption{ Accuracy of the proposed heuristic approach with Coeffecient (a) 0.15 (b) 0.20 (c)  0.25 (d) Accuracy of the greedy approach (e) Accuracy of the heuristic and greedy approach (f) Execution time in log scale }
  \label{fig_sim}
  \vspace{-5mm}
\end{figure}

\subsection{Accuracy Analysis}
We have applied the proposed heuristic approach to select and compose relatively similar consumers according to the provider’s CP-Net. The accuracy of the heuristic approach is calculated with respect to the brute force approach. We also compose the consumers based on the greedy selection approach. The accuracy of the greedy approach is calculated in a similar manner.

The brute force approach normally provides more accurate result then the heuristic approach as the brute force approach considers all possible combination of consumers. For the heuristic approach, we select a consumer in the composition only if its CP-Net is more than 20\% relatively similar to the provider’s CP-Net. We have run the experiments several times to find the optimal threshold. For a specific provider and a service, this threshold should be set manually before composition. Figure \ref{fig_sim}(a), (b), and (c) show the histogram of the accuracy of the heuristic approach where the correlation coefficients are 0.15, 0.20, and 0.25. The brute force approach provides the optimal results. Compared to the brute force result, the heuristic approach generates almost 60\% accurate result on average when the coefficient is 0.2. Figure \ref{fig_sim}(d) shows the accuracy result of the greedy based approach. The result shows the histogram of the outcomes. Here, outcomes have below 50\% accuracy most of the time. Figure \ref{fig_sim} (e) shows the average accuracy of the proposed heuristic approach and the greedy approach. The greedy approach provides very good accuracy if the number of consumers is low. The accuracy of the greedy approach becomes low with the increase of consumers, as it starts to discard more consumers. The heuristic approach provides better accuracy with the increase in the number of consumers because it finds more similar consumers according to the provider’s CP-Net.

\subsection{Runtime Analysis}
Figure \ref{fig_sim}(f) depicts the time comparison between the brute force, heuristic, and greedy approaches in log scale. The figure shows that with the increase in the number of consumers, time for composition in the brute force approach grows exponentially ( i.e., linearly in log scale). The heuristic approach does not show exponential behavior. For a particular IaaS provider, the composition time increases very slowly compared to the brute force approach. The greedy approach shows a very interesting result. It composes consumers with a constant time for a particular service. The accuracy of the greedy approach is unreliable.

\section{Related Work}

Qualitative user-preferences are represented by graphical models, especially in multi-objective decision-making domain \cite{lang2009sequential}.  IaaS consumers can express their preferences more directly, intuitively using qualitative representations. A conditional preference network (CP-Net) provides a natural and an efficient way to represent consumers’ preferences in qualitative manner \cite{boutilier2004cp}. CP-Nets are widely used to represent user’s preferences to select and compose services \cite{wang2017incorporating}. A web service selection mechanism is proposed to incorporate incomplete or inconsistent user preferences from historical preferences \cite{wang2017incorporating}. A CP-Net based similarity measurement approach is proposed to find users with similar preferences \cite{wang2017measuring}. Several variations of CP-Nets are also proposed to enhance the expressiveness of users. A UCP-Net is a graphical representation of utility functions that combines generalize additive models and CP-Nets \cite{boutilier2001ucp}. The TCP-Net is another variation of CP-Net where relative importance between attributes can be captured through weighted edges \cite{mukhtar2009quantitative}. The WCP-Net is another weighted CP-Net that is proposed to capture user’s preference more precisely to select web services. A deterministic temporal CP-Net is used to express IaaS provider’s long-term business strategies \cite{mistry2016qualitative}. A probabilistic CP-Net is proposed to capture the IaaS provider’s business strategies in a probabilistic manner \cite{mistry2017probabilistic}.  

Several existing studies propose different methods to compose multiple CP-Nets. A multi-agent CP-Net or mCP-Net is proposed as an extension of CP-Net \cite{rossi2004mcp}. Preferences from multiple users are aggregated into a single CP-Net. The proposed method aggregates preferences according to a voting analogy where preferences are selected based on the preference of majority user. A Majority-rule-based preference aggregation method is proposed based on a hypercube-wise composition to optimize the composition process. An aggregation method is proposed to capture multi-valued CP-Nets based on majoritarian aggregation rule \cite{li2015aggregating}. The proposed method can aggregate CP-Nets even if they are cyclic. Most existing works on the aggregation of CP-Nets consider the composition problem as a multi-agent voting system. These approaches are not applicable in the case of resource allocation based on multi-agent preferences. We compose CP-Nets based on the  composability model and the resource constraints in our proposed composition approach. The composed CP-Nets capture the preferences of multi-users instead of just considering the common preferences.

\section{Conclusion}
We propose a CP-Net based composition approach for an IaaS provider. The proposed approach allows the IaaS provider and consumers to express their qualitative conditional preferences using CP-Nets. We propose a composability model for IaaS consumers using the semantic congruence property of a qualitative composition. Finding the optimal composition may be difficult when the number of consumers is large. A greedy-based and a heuristic-based selection approaches are proposed to reduce the search space of candidate consumers. Both approaches utilize correlation coefficients between CP-Nets to find consumers who have similar preferences with the provider. Experimental results show that the proposed heuristic-based approach is applicable to the runtime and the performance is acceptable. One key limitation of the proposed approach is that it considers only the deterministic model of CP-Nets. In the future, we want to explore the composability model of probabilistic CP-Nets in the context of the long-term IaaS composition.
\section{Acknowledgement}
This research was made possible by NPRP 7-481-1-088 grant from the Qatar National Research Fund (a member of The Qatar Foundation). The statements made herein are solely the responsibility of the authors.

%
%
%

 \bibliographystyle{splncs04}
 \bibliography{references}
\end{document}